\documentclass[journal]{IEEEtran}

\usepackage[T1]{fontenc}
\usepackage{amsmath,amssymb,amsfonts}
\usepackage{algorithmic}
\usepackage{algorithm}
\usepackage{graphicx}
\usepackage{textcomp}
\usepackage{xcolor}
\usepackage{booktabs}
\usepackage{array}
\usepackage{multirow}
\usepackage{url}
\usepackage{tikz}
\usetikzlibrary{shapes.geometric,arrows.meta,positioning,calc,fit,backgrounds}
\usepackage[hidelinks]{hyperref}
\usepackage[capitalize,noabbrev]{cleveref}

\graphicspath{{figures/}}

\def\BibTeX{{\normalfont B\kern-.05em{\scshape i\kern-.025em b}\kern-.08em
    T\kern-.1667em\lower.7ex\hbox{E}\kern-.125emX}}

\newcommand{\pistar}{\pi^{\star}}
\newcommand{\pizero}{\pi_{0}}
\newcommand{\Aop}{\mathcal{A}}
\newcommand{\Lampii}{\textsc{LaMP-2}}
\newcommand{\Lampiii}{\textsc{LaMP-3}}
\newcommand{\ns}{\textsc{ns}}

\begin{document}

\title{Prompt-Space Meta-Learning Does Not Transfer Across Users: A Frozen-LLM Negative Result}

\author{Liam~Byrne,~David~Dylan,~Orla~Fitzgerald,~Eoin~Doyle,~Ciara~Nolan,~Padraig~Lynch,~and~Sinead~Gallagher
\thanks{D.~Dylan, L.~Byrne, C.~Nolan and S.~Gallagher are with Trinity College Dublin; O.~Fitzgerald and P.~Lynch are with University College Dublin; E.~Doyle is with Dublin City University, Dublin, Ireland.}%
}

\markboth{IEEE Transactions on Knowledge and Data Engineering}%
{Byrne \MakeLowercase{\textit{et al.}}: Meta-Objective Collapse in Prompt-Space Meta-Learning}

\maketitle

\begin{abstract}
Personalizing a frozen large language model (LLM) to individual users is
often framed as a meta-learning problem in prompt space: each user is a
task, and one seeks a shared natural-language adaptation policy that, given a
handful of the user's labeled interactions, configures the frozen model for that
user. The framing is attractive because it is backbone-agnostic and reuses the
machinery of prompt optimization, yet the field rarely tests whether the
optimized meta-objective encodes transferable cross-user adaptation rather than
generic instruction quality. We study this question with \textsc{Muse}
(Meta-learned User-adaptation via Shared Evolution), which evolves a single
shared adaptation prompt over a meta-train user population by reflective prompt
evolution, freezes it, and applies it zero-shot to held-out users; matched
controls isolate learning from confounds of phrasing and selection. On two
standard personalization benchmarks (\textsc{LaMP-2} categorization and
\textsc{LaMP-3} rating) over 200 held-out users each, \textsc{Muse} does not
significantly improve on its own un-evolved seed prompt or on a structure-broken
control that meta-trains on mismatched user--support pairs, and is dominated by
plain few-shot retrieval on the rating task ($\Delta$MAE $+0.175$, $p<0.001$). We attribute these outcomes to a single
mechanism, \emph{meta-objective collapse}: the meta-validation objective is
statistically invariant to whether the user--support correspondence is genuine
($p{=}0.555$ on \textsc{LaMP-2}, $p{=}0.622$ on \textsc{LaMP-3}), so it cannot
be optimized into transferable adaptation and instead rewards instruction polish
and validation overfitting. The seed-prompt, wrong-support, and
invariance--oracle controls form a reusable protocol that separates learned
adaptation from these confounds.
\end{abstract}

\begin{IEEEkeywords}
Few-shot learning, frozen language models, large language models,
meta-learning, negative transfer, personalization, prompt optimization,
transfer learning, user modeling.
\end{IEEEkeywords}

\IEEEpeerreviewmaketitle

\begin{figure*}[t]
  \centering
  \IfFileExists{figures/fig_teaser.pdf}{%
    \includegraphics[width=\textwidth]{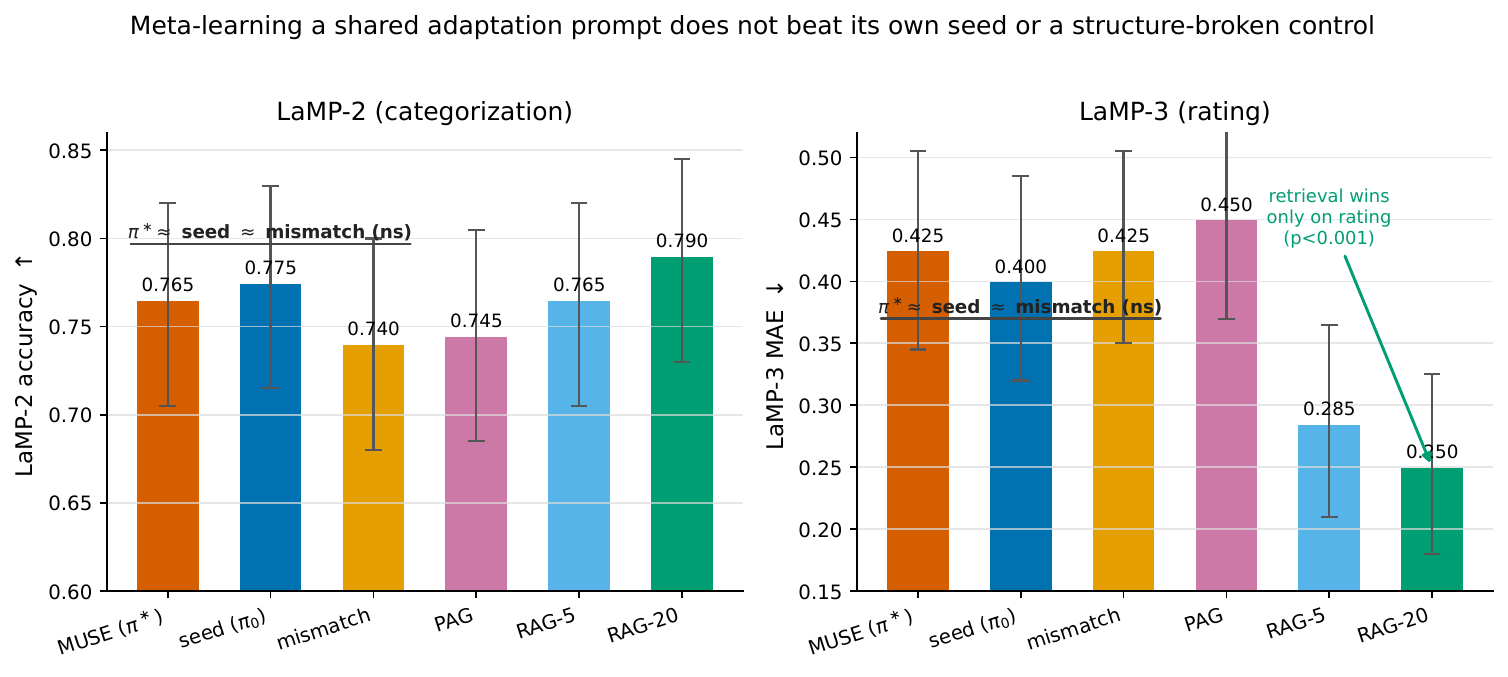}%
  }{%
    \fbox{\parbox[c][4.5cm][c]{0.95\textwidth}{\centering
      \textbf{[ teaser placeholder ]}\\[2pt]
      \texttt{figures/fig\_teaser.pdf} will be inserted here.}}%
  }
  \caption{The headline negative result. On both benchmarks, meta-learning a
  shared adaptation prompt with \textsc{Muse} ($\pi^{\star}$) is
  statistically indistinguishable from applying its own un-evolved seed prompt
  ($\pi_{0}$) and from a structure-broken control whose persona policy was
  meta-trained on deliberately mismatched user--support pairs (\emph{mismatch}).
  All three also tie the strongest persona baseline (\textsc{Pag}). On the
  \textsc{LaMP-3} rating task, plain few-shot retrieval (\textsc{Rag}) wins
  decisively ($p<0.001$); on \textsc{LaMP-2} categorization all methods cluster
  near parity. Bars are accuracy ($\uparrow$, left) and mean absolute error
  ($\downarrow$, right) over $200$ held-out users; whiskers are $95\%$
  bootstrap confidence intervals. The phrase ``meta-learning does not beat its
  own seed or a structure-broken control'' is the thesis of this paper.}
  \label{fig:teaser}
\end{figure*}

\section{Introduction}\label{sec:intro}

\IEEEPARstart{P}{ersonalization} has become a central requirement for large
language model (LLM) services: the same model is expected to serve millions of
users, yet each user brings idiosyncratic preferences, labeling tendencies, and
style. Because retraining or even parameter-efficient finetuning a hosted model
\emph{per user} is operationally infeasible at scale, a large and growing body
of work personalizes a \emph{frozen} backbone by manipulating its
input---retrieving the user's past interactions, distilling them into a compact
persona, or prepending a learned soft or natural-language
prompt~\cite{salemi2024lamp,richardson2023pag,zhang2025personaagent,tan2024oppu,zhang2024personalizationsurvey}.
A natural and attractive way to systematize this is to view it as a
\emph{meta-learning} problem. If we treat each user as a task, with the user's
labeled history as a support set and their future interactions as queries, then
personalization is exactly the few-shot, fast-adaptation regime that
meta-learning was designed for~\cite{finn2017maml,hospedales2021survey}. One
would like to meta-learn, across a population of users, a single
\emph{adaptation policy} that maps any new user's support set to a configuration
of the frozen model that performs well on that user's queries. Because the
backbone is frozen and only its textual input is mutable, the most direct
instantiation of this idea lives in \emph{prompt space}: meta-learn a shared
natural-language ``adaptation prompt'' that, when combined with a user's
history, induces a per-user persona. This framing is appealing on several
counts---it is backbone-agnostic, it reuses the mature machinery of prompt
optimization and evolution~\cite{zhou2023ape,yang2024opro,agrawal2025gepa}, and
it promises amortized, transferable adaptation rather than a bespoke pipeline
per user.

This paper asks whether prompt-space meta-learning actually learns transferable
cross-user structure, or whether it merely appears to. We address the question
with a clean instantiation of the idea, \textbf{\textsc{Muse}} (Meta-learned
User-adaptation via Shared Evolution), in which a single shared adaptation prompt
$\pi$ is evolved over a meta-train population of users using reflective prompt
evolution in the style of \textsc{Gepa}~\cite{agrawal2025gepa}, scored by each
candidate's accuracy (or negative error) on meta-train users' own held-out
queries, selected on a \emph{disjoint} meta-validation population, frozen as
$\pistar$, and applied zero-shot to build a persona for each of $200$ held-out
test users. \textsc{Muse} sits in an apples-to-apples harness with a frozen
Qwen3-30B-A3B backbone~\cite{yang2025qwen3}, deterministic benchmark scoring, and
a battery of strong baselines (no-personalization, random-user, profile-augmented
generation, distilled personas, and retrieval-augmented few-shot at several
budgets). To our knowledge, this is the first controlled head-to-head test of
\textsc{Gepa}-style prompt evolution \emph{as cross-user meta-learning} for
frozen-LLM personalization.

The findings are consistent and, at first sight, counterintuitive. \textsc{Muse}
with the evolved $\pistar$ does \emph{not} significantly outperform the same
pipeline run with its hand-written seed prompt $\pizero$; the seed is in fact
numerically better on both benchmarks. \textsc{Muse} also does \emph{not} beat a
deliberately structure-broken control in which the shared prompt is meta-trained
on \emph{mismatched} user--support pairs, where each user's persona is built from
a different user's history, and it is statistically tied with the strongest
persona baselines. On the rating benchmark, simple few-shot retrieval dominates
all persona methods by a wide and highly significant margin. These are not
isolated null results but symptoms of a single mechanism, which we name and
quantify as \textbf{meta-objective collapse}. The aggregate meta-validation
objective that the optimizer climbs is structure-blind: it is statistically
invariant to whether the user--support correspondence is genuine. Deranging the
user-to-support assignment during meta-training, which destroys the adaptive
signal by construction, changes the selection objective by an amount
indistinguishable from zero. An objective that cannot distinguish aligned from
scrambled supports cannot, even in principle, be optimized into transferable
adaptation; the optimizer can acquire only generic instruction polish, which the
seed prompt already supplies, together with overfitting of a small ($30$-user)
validation population. This is a concrete instance of \emph{proxy-objective
misalignment}: the meta-validation proxy that is optimized is decoupled from the
true goal of transferable per-user adaptation.

This is a negative result, and we report it as one: the headline numbers, the
retrieval advantage on regression, and every confidence interval are stated
directly. The contribution lies in the rigor of the diagnosis and in the controls
that produce it. The same three controls, namely a seed-prompt baseline, a
wrong-support derangement, and an invariance/oracle decomposition, constitute a
reusable evaluation protocol that can be applied before attributing learned
adaptation to a prompt-space method. Without them, the instruction polish and
validation overfit that \textsc{Muse} exhibits are easily mistaken for genuine
cross-user transfer. \Cref{fig:teaser} previews the finding.

\smallskip
\noindent\textbf{Contributions.} The main contributions of this paper are
summarized as follows.
\begin{itemize}
  \item \textbf{\textsc{Muse}: a clean instantiation and test of prompt-space
  meta-learning for personalization.} We formalize users-as-tasks
  personalization of a frozen LLM and instantiate it with \textsc{Muse}, the
  first apples-to-apples test of \textsc{Gepa}-style reflective prompt evolution
  as a cross-user meta-learner. \textsc{Muse} evolves one shared
  natural-language adaptation prompt over a meta-train user population, freezes
  it, and applies it zero-shot to held-out users
  (\cref{sec:problem,sec:method}).
  \item \textbf{The meta-objective collapse diagnosis via a derangement
  control.} Using a mismatch (wrong-support) control, we show that the
  meta-validation objective is \emph{statistically invariant} to scrambling the
  user--support correspondence ($\Delta$ accuracy $+0.033$, $95\%$ CI
  $[-0.050,+0.117]$, $p{=}0.555$ on \Lampii{}; $\Delta$ reward $+0.013$,
  $[-0.029,+0.054]$, $p{=}0.622$ on \Lampiii{}). An objective blind to this
  structure carries no exploitable cross-user adaptive signal
  (\cref{sec:analysis}).
  \item \textbf{Quantified meta-overfitting at small population scale.} We
  measure a uniform optimistic gap of $\approx\!9$ points between
  meta-validation and meta-test on \Lampii{}, show that meta-validation does not
  significantly rank held-out test performance (Spearman $\rho{=}+0.47$,
  $p{=}0.20$, \ns{}), and observe flat, non-monotone support-size and
  meta-train-size sweeps---the signature of selection-set overfitting rather than
  learned adaptation (\cref{sec:analysis}).
  \item \textbf{An invariance/oracle decomposition and a reusable evaluation
  protocol.} We decompose test behavior into phrasing-invariant and
  phrasing-sensitive cases ($80\%$/$72\%$ invariant on \Lampii{}/\Lampiii{}),
  attribute the oracle headroom to phrasing variance rather than learned
  adaptation, and package the seed-prompt, wrong-support, and invariance/oracle
  controls into a protocol for distinguishing learned adaptation from
  instruction polish and validation overfit (\cref{sec:analysis,sec:discussion}).
\end{itemize}

\smallskip
\noindent\textbf{Roadmap.} \Cref{sec:related} surveys LLM personalization,
meta-learning and fast adaptation, in-context learning as meta-learning, prompt
optimization, test-time adaptation, and retrieval, and positions our study on
the \emph{learning} axis (does the meta-objective improve adaptation over a
seed?), which is orthogonal to the representation axis of distillation versus
retrieval. \Cref{sec:problem} formalizes users-as-tasks personalization.
\Cref{sec:method} details \textsc{Muse}. \Cref{sec:setup} describes the
benchmarks, baselines, backbone, metrics, and controls. \Cref{sec:results}
reports the main results. \Cref{sec:analysis} establishes meta-objective
collapse through its three mechanistic legs plus the invariance/oracle
decomposition. \Cref{sec:discussion} draws the protocol-level lessons, states
when prompt-space meta-learning can and cannot be expected to work, and lists
limitations. \Cref{sec:conclusion} concludes.

\section{Related Work}\label{sec:related}

We organize prior work along six threads that jointly define the setting and the
methods we test: (i)~LLM personalization and user modeling; (ii)~meta-learning
and fast adaptation; (iii)~in-context learning understood as meta-learning;
(iv)~prompt optimization and evolution; (v)~test-time adaptation; and
(vi)~retrieval and retrieval-augmented generation. Throughout, we distinguish
two orthogonal axes. The \emph{representation} axis asks how a user's history is
encoded into the frozen model's input (e.g., a distilled persona versus
retrieved exemplars); the sibling study ``Distill or Retrieve?'' isolates that
axis and we therefore reference it exactly once, below, to make clear that our
contribution is on a different axis. The \emph{learning} axis---our focus---asks
whether a meta-objective that climbs over a user population improves adaptation
beyond a fixed seed policy. Our negative finding lives entirely on the learning
axis.

\subsection{LLM Personalization and User Modeling}
The Language Model Personalization (LaMP) benchmark suite established a standard
protocol in which each user supplies a profile of past, labeled interactions and
the model must predict the user's behavior on new
items~\cite{salemi2024lamp,kumar2024longlamp}. A dominant family of methods
augments the prompt with the user's profile, either directly or after
summarization. Profile-Augmented Generation (PAG) distills a user's history into
a natural-language profile with a fixed, hand-written
instruction~\cite{richardson2023pag}; retrieval-of-profile-then-generate
pipelines select the most relevant past items before
answering~\cite{salemi2024ropg}. PersonaAgent and related agentic personalizers
iterate a persona through self-feedback~\cite{zhang2025personaagent}, while
plug-in user encoders such as P-PLUG and \textsc{UserLLM} compress a user into
soft tokens or embeddings that condition the
backbone~\cite{liu2025pplug,ning2024userllm}. Parameter-side personalization,
exemplified by OPPU and per-user parameter-efficient
modules~\cite{tan2024oppu,tan2024perpcs,han2023personapkt}, finetunes a small
adapter per user; these are powerful but assume write access to model weights,
which the hosted-frozen-backbone setting precludes. A complementary literature
models users for recommendation and preference
prediction~\cite{wu2023recsurvey,zhang2023malp,mysore2023pearl}, builds explicit
long-term memory of user state~\cite{packer2023memgpt,zhong2024memorybank}, and
studies persona consistency and role
fidelity~\cite{zhang2018personachat,wolf2019transfertransfo,tseng2024persona,shao2023characterllm,wang2024rolellm,tu2024charactereval,samuel2024personagym,chen2024rplasurvey}.
Surveys of LLM personalization and of personalized
LLMs~\cite{zhang2024personalizationsurvey,liu2025pllmsurvey,wang2023cuecot}
catalog these approaches but, as we argue here, rarely separate genuine learned
adaptation from prompt-engineering artifacts. Our \textsc{Muse} uses the same
distilled-persona representation as PAG and PersonaLink, but replaces their
fixed adaptation instruction with a meta-learned one, isolating exactly the
``does meta-learning help?'' question.

\subsection{Meta-Learning and Fast Adaptation}
Meta-learning seeks an inductive bias---an initialization, an optimizer, or a
metric space---that enables rapid adaptation to new tasks from few
examples~\cite{hospedales2021survey}. Gradient-based methods learn an
initialization from which a few SGD steps suffice: MAML and its
relatives~\cite{finn2017maml,nichol2018reptile,li2017metasgd,raghu2020anil,antoniou2019howtotrain},
and learned-optimizer approaches~\cite{andrychowicz2016l2l,ravi2017optasfewshot}.
Metric-based methods learn an embedding in which simple nearest-neighbor or
prototype rules
generalize~\cite{snell2017protonet,vinyals2016matchingnet,oreshkin2018tadam}, and
memory- or attention-augmented learners amortize adaptation in a forward
pass~\cite{santoro2016mann,mishra2018snail}. Continual and online variants study
adaptation without catastrophic
forgetting~\cite{javed2019oml}, and the paradigm has been applied to
recommendation, machine translation, and control as a few-shot user/task
problem~\cite{lee2019melu,gu2018metanmt,yu2019metaworld}. The users-as-tasks view
we adopt is the natural transcription of this literature to personalization: a
user is a task, the profile is a support set, and we seek a shared policy that
adapts fast. A central and well-known risk in meta-learning is that the
meta-objective can be optimized in ways that do not produce genuine
task-adaptive behavior---``memorization'' or shortcut solutions that exploit the
meta-training distribution rather than learning to
adapt~\cite{raghu2020anil,yu2019metaworld}. Our negative result is a sharp,
prompt-space manifestation of exactly this risk.

\subsection{In-Context Learning as Meta-Learning}
A parallel line interprets in-context learning (ICL) itself as implicit
meta-learning: a model trained on many sequences learns to infer a latent task
from a few in-context examples and apply it to a query. \textsc{MetaICL}
explicitly meta-trains a model to do ICL across
tasks~\cite{min2022metaicl}; theoretical and mechanistic analyses show that
transformers can implement learning algorithms (e.g., gradient descent or ridge
regression) in their forward pass over the
context~\cite{akyurek2023icllinear,vonoswald2023icgd}. This perspective is
important for our study because it clarifies what a frozen backbone already does:
given a user's exemplars in context, the model performs a form of implicit
adaptation \emph{for free}. The retrieval baselines we compare against exploit
precisely this---they place raw user exemplars in the context and let the frozen
model adapt implicitly---whereas \textsc{Muse} attempts to meta-learn an
\emph{explicit} adaptation instruction on top. One reading of our results is that
the explicit prompt-space meta-objective adds little over the implicit
adaptation the backbone already supplies.

\subsection{Prompt Optimization and Evolution}
Because the backbone is frozen, prompt-space personalization inherits the
toolkit of automatic prompt optimization. Discrete search and reinforcement
methods such as APE and OPRO search instruction text against a validation
metric~\cite{zhou2023ape,yang2024opro}; \textsc{DSPy} compiles and optimizes
multi-stage prompt programs~\cite{khattab2024dspy}; and evolutionary or
reflective methods---Promptbreeder, ExpeL, agentic memory induction, and
notably \textsc{Gepa}'s genetic--Pareto reflective
evolution---mutate prompts from natural-language feedback on prior
rollouts~\cite{fernando2023promptbreeder,zhao2024expel,wang2024awm,agrawal2025gepa,ace2025}.
Reflective self-improvement of model outputs more
broadly~\cite{madaan2023selfrefine,shinn2023reflexion,gou2023critic,huang2022selfimprove,yuan2024selfrewarding,wu2024metarewarding}
shares the same engine: an LLM critiques and rewrites text against a target. Our
\textsc{Muse} uses \textsc{Gepa}'s loop verbatim, but elevates the optimized
prompt from a per-task strategy to a \emph{shared, frozen, cross-user adaptation
policy}. This is what makes it a meta-learner rather than a per-instance
optimizer, and it is also what exposes the failure mode: when the optimization
target is an aggregate over a user population, the target can be structure-blind
even when each per-user evaluation looks reasonable. We frame this failure as
\emph{proxy-objective misalignment}, in the sense surveyed for LLM and agent
objectives by~\cite{yang2026misalignment}: the meta-validation proxy that
\textsc{Gepa} maximizes is decoupled from the true goal of transferable
adaptation, and our derangement control \emph{is} the operational demonstration
of that proxy-versus-goal gap.

\subsection{Test-Time and Grounded Adaptation}
Test-time adaptation updates a model on the fly using the test input itself.
Test-time training and entropy-minimization methods such as TTT, Tent, and
gradient-edited variants adapt parameters per
instance~\cite{sun2020ttt,wang2021tent,mitchell2022mend}, while recent ``thinking
at test time'' approaches scale inference-time computation against a verifiable
signal~\cite{snell2024testtime,muennighoff2025s1,yao2023tot,wang2023selfconsistency}.
What unites the \emph{successful} members of this family is a per-instance,
grounded feedback signal: a self-consistency check, a verifier, a unit test, or
a tool-execution result that tells the adapter, for \emph{this} input, whether it
is on track. The contrast with our setting is the crux of the paper. Grounded,
instance-level test-time adaptation---for example, agentic tool reasoning that is
verified per call against execution feedback~\cite{yang2026tooltree}---can
genuinely improve because its objective is tied to a checkable outcome on each
instance. \textsc{Muse}'s meta-objective is the opposite: it is an aggregate over
a population, with no per-instance grounding, and that is precisely why it
collapses. What \textsc{Muse} lacks is exactly the instance-level, verifiable
feedback that makes grounded test-time adaptation work. We return to this
contrast in \cref{sec:discussion}.

\subsection{Retrieval and Retrieval-Augmented Generation}
The strongest baselines in our study are retrieval-augmented few-shot prompts,
which fetch a user's most relevant past items and place them, verbatim, in the
context. This builds on classical sparse and dense
retrieval~\cite{robertson2009bm25,karpukhin2020dpr,izacard2021contriever,xiong2021ance,formal2021splade,khattab2020colbert},
retrieval-augmented language modeling and
generation~\cite{lewis2020rag,guu2020realm,borgeaud2022retro,izacard2021fid,ram2023incontextralm,shi2024replug,khattab2022dsp,asai2024selfrag},
and demonstration selection for
ICL~\cite{liu2022goodexamples,rubin2022epr,wang2024llmretriever}. For
personalization specifically, retrieving the user's own history is a very strong,
training-free baseline because it preserves fine-grained, instance-level signal
(e.g., a user's exact rating calibration) that a distilled persona discards. The
orthogonal axis of \emph{representation}---whether to distill a user into a
compact persona or to retrieve raw exemplars---is studied directly by the sibling
work ``Distill or Retrieve?''; we cite it here once to delimit scope and to make
explicit that the present paper does \emph{not} re-litigate that axis. Our study
holds the representation fixed (a distilled persona, as in PAG/PersonaLink) and
varies only the \emph{learning} of the adaptation policy, asking whether
meta-learning the policy beats a seed. The retrieval baselines enter our tables
as a reference point and, on the regression task, as the decisive winner---an
honest fact we report rather than minimize.

\section{Problem Formulation}\label{sec:problem}

\subsection{Users as Tasks}
Let $f$ denote a \emph{frozen} autoregressive LLM that maps a textual prompt to a
textual output; its parameters are fixed and never updated. Personalization must
therefore act entirely through $f$'s input. We adopt an episodic, users-as-tasks
formulation. A user $u$ is a task with an associated distribution
$\mathcal{D}_u$ over labeled items $(x,y)$, where $x$ is an input (e.g., a news
article or a product review) and $y$ is the label this user would assign (a
category or a rating). Each user exposes a finite \emph{support set}
$S_u=\{(x_i,y_i)\}_{i=1}^{m_u}$ (their profile of past labeled interactions) and
is evaluated on held-out \emph{query} items drawn from the same $\mathcal{D}_u$.
The population of users is partitioned into three mutually disjoint sets: a
meta-train set $\mathcal{U}_{\text{tr}}$, a meta-validation set
$\mathcal{U}_{\text{val}}$, and a meta-test set $\mathcal{U}_{\text{te}}$. No
user appears in more than one set, so generalization is measured across users,
not merely across items within a user. \Cref{tab:notation} collects the
notation.

\begin{table}[t]
\centering
\caption{Notation used throughout the paper.}
\label{tab:notation}
\footnotesize
\renewcommand{\arraystretch}{1.18}
\begin{tabular}{@{}c l@{}}
\toprule
\textbf{Symbol} & \textbf{Meaning} \\
\midrule
$f$ & frozen backbone LLM (greedy decoding, parameters fixed) \\
$u$ & a user, identified with a meta-task \\
$\mathcal{D}_u$ & user $u$'s item distribution over $(x,y)$ \\
$S_u$ & user $u$'s support set (labeled profile) \\
$Q_u$ & user $u$'s held-out query items \\
$\mathcal{U}_{\text{tr}},\mathcal{U}_{\text{val}},\mathcal{U}_{\text{te}}$ &
 disjoint meta-train / meta-val / meta-test user sets \\
$\pi$ & shared natural-language adaptation prompt (the meta-parameter) \\
$\pizero$ & hand-written seed adaptation prompt \\
$\pistar$ & evolved, frozen adaptation prompt selected on $\mathcal{U}_{\text{val}}$ \\
$\Aop(\pi,S_u)$ & inner adaptation: build user $u$'s persona under $\pi$ \\
$\theta_u$ & the induced per-user persona (preamble text) \\
$\ell$ & item-level loss / score (accuracy or absolute error) \\
$\mathcal{M}(\pi)$ & meta-objective: population score of $\pi$ \\
$\sigma$ & a derangement of the user$\to$support assignment \\
$K$ & support-set cap (items per user used to build the persona) \\
$\rho,\tau$ & Spearman / Kendall rank correlations \\
\bottomrule
\end{tabular}
\end{table}

\subsection{Inner Adaptation and the Meta-Objective}
Adaptation to a user is mediated by a shared, learnable natural-language
\emph{adaptation prompt} $\pi$, the meta-parameter. The inner adaptation operator
$\Aop$ takes $\pi$ and a user's support set and produces a per-user persona
$\theta_u$ by a single call to the frozen model,
\begin{equation}
\theta_u \;=\; \Aop(\pi, S_u) \;=\; f\big(\,\textsc{distill}(\pi, S_u)\,\big),
\label{eq:inner}
\end{equation}
where $\textsc{distill}(\pi,S_u)$ is a fixed template that injects $\pi$ as the
controlling instruction over a digest of the user's history and elicits a bounded
persona (a preference summary, a few exemplars, and a few decision rules; see
\cref{sec:method}). The persona $\theta_u$ is then prepended to a query and the
frozen model answers,
\begin{equation}
\hat{y} \;=\; f\big(\,\theta_u \,\Vert\, x\,\big), \qquad (x,y)\in Q_u,
\label{eq:answer}
\end{equation}
with ``$\Vert$'' denoting prompt concatenation. The per-user performance of $\pi$
is the expected item score on that user's queries,
\begin{equation}
J_u(\pi) \;=\; \mathbb{E}_{(x,y)\sim Q_u}\Big[\,\ell\big(f(\Aop(\pi,S_u)\Vert x),\,y\big)\Big],
\label{eq:peruser}
\end{equation}
where $\ell$ is accuracy for categorization and a bounded negative-error reward
for rating (so that larger is better in both cases). The \emph{meta-objective} is
the population average over a user set $\mathcal{U}$,
\begin{equation}
\mathcal{M}_{\mathcal{U}}(\pi) \;=\; \frac{1}{|\mathcal{U}|}\sum_{u\in\mathcal{U}} J_u(\pi).
\label{eq:meta}
\end{equation}
Meta-learning seeks $\pistar = \arg\max_{\pi}\,\mathcal{M}_{\mathcal{U}_{\text{tr}}}(\pi)$,
with the final $\pistar$ \emph{selected} by its score
$\mathcal{M}_{\mathcal{U}_{\text{val}}}$ on the disjoint meta-validation users,
and reported on the held-out meta-test users via the benchmark's native metric.

\subsection{What ``Transfer'' Requires, and the Structure-Blindness Test}
For prompt-space meta-learning to be meaningful, $\mathcal{M}_{\mathcal{U}}(\pi)$
must reward $\pi$ for producing personas that are genuinely \emph{adapted} to the
right user---that is, the objective must depend on the correspondence between a
user's support set $S_u$ and that user's queries $Q_u$. We make this requirement
operational with a derangement. Let $\sigma$ be a permutation of
$\mathcal{U}_{\text{tr}}$ with no fixed point, and define the
\emph{mismatched} meta-objective
\begin{equation}
\mathcal{M}^{\sigma}_{\mathcal{U}}(\pi) \;=\; \frac{1}{|\mathcal{U}|}\sum_{u\in\mathcal{U}}
\mathbb{E}_{(x,y)\sim Q_u}\Big[\,\ell\big(f(\Aop(\pi,S_{\sigma(u)})\Vert x),\,y\big)\Big],
\label{eq:metashuf}
\end{equation}
in which each user's persona is built from \emph{another} user's support set
$S_{\sigma(u)}$ but still graded on the original user's queries $Q_u$. If the
objective carries genuine cross-user adaptive signal, then aligned supports must
help relative to scrambled ones, i.e.,
$\mathcal{M}_{\mathcal{U}}(\pi) > \mathcal{M}^{\sigma}_{\mathcal{U}}(\pi)$ by a
margin the optimizer can exploit. We define \textbf{meta-objective collapse} as
the empirical condition
\begin{equation}
\mathcal{M}_{\mathcal{U}_{\text{val}}}(\pi) \;\approx\; \mathcal{M}^{\sigma}_{\mathcal{U}_{\text{val}}}(\pi),
\label{eq:collapse}
\end{equation}
i.e., the meta-objective is statistically invariant to whether the
user--support correspondence is real. Under \cref{eq:collapse}, no optimizer can
convert $\mathcal{M}$ into transferable adaptation, because the quantity it
climbs does not distinguish adaptation from its structure-destroyed counterpart.
\Cref{sec:analysis} shows that \cref{eq:collapse} holds on both benchmarks.

\section{Method: \textsc{Muse}}\label{sec:method}

\begin{figure*}[t]
\centering
\footnotesize
\begin{tikzpicture}[
  node distance=6mm and 9mm,
  box/.style={draw, rounded corners=2pt, align=center, inner sep=4pt,
    minimum height=8mm, font=\footnotesize},
  data/.style={box, fill=blue!6},
  proc/.style={box, fill=orange!10},
  frozen/.style={box, fill=gray!12},
  outbox/.style={box, fill=green!8},
  >={Stealth[length=2mm]},
  every path/.style={thick}
]
\node[data] (tr) {Meta-train users\\$\mathcal{U}_{\text{tr}}$ (40)\\
  \scriptsize support $S_u$ / query $Q_u$};
\node[proc, right=of tr] (gepa) {\textsc{Gepa} evolve\\(reflective mutation\\
  of shared $\pi$)};
\node[proc, below=5.5mm of gepa] (score) {score $\pi$ on each\\user's held-out
  $Q_u$\\\scriptsize \cref{eq:peruser}--\eqref{eq:meta}};
\node[data, below=5.5mm of tr] (val) {Meta-val users\\$\mathcal{U}_{\text{val}}$
  (30, disjoint)};
\node[frozen, right=14mm of gepa] (pistar) {Freeze\\$\pi^{\star}$};
\node[data, right=12mm of pistar] (te) {Held-out user\\$u\in\mathcal{U}_{\text{te}}$
  (200)\\\scriptsize support $S_u$};
\node[proc, below=5.5mm of te] (persona) {Distill persona\\
  $\theta_u=\mathcal{A}(\pi^{\star},S_u)$\\\scriptsize 1 frozen-LLM call};
\node[outbox, below=5.5mm of persona] (ans) {Answer query\\
  $\hat{y}=f(\theta_u\,\Vert\,x)$\\\scriptsize 1 frozen-LLM call};
\draw[->] (tr) -- (gepa);
\draw[->] (gepa) to[bend left=12] (score);
\draw[->] (score) to[bend left=12] node[right,font=\scriptsize]{feedback} (gepa);
\draw[->] (val) -- node[below,font=\scriptsize,pos=0.45]{select best $\pi$} (score.south west);
\draw[->] (gepa) -- node[above,font=\scriptsize]{$\arg\max\mathcal{M}_{\mathcal{U}_{\text{val}}}$} (pistar);
\draw[->] (pistar) -- node[above,font=\scriptsize]{zero-shot} (te);
\draw[->] (te) -- (persona);
\draw[->] (persona) -- (ans);
\draw[->] (pistar.south) to[bend right=18] (persona.north west);
\begin{scope}[on background layer]
  \node[draw=blue!40, dashed, rounded corners, fit=(tr)(val)(gepa)(score),
    inner sep=5pt, label={[font=\scriptsize\itshape,blue!50!black]above:meta-train (evolve $\pi$)}] {};
  \node[draw=green!45!black, dashed, rounded corners, fit=(te)(persona)(ans),
    inner sep=5pt, label={[font=\scriptsize\itshape,green!40!black]above:meta-test (frozen $\pi^{\star}$, zero-shot)}] {};
\end{scope}
\end{tikzpicture}
\caption{The \textsc{Muse} pipeline. A single shared adaptation prompt $\pi$ is
evolved over a meta-train user population with \textsc{Gepa}-style reflective
mutation, where each candidate $\pi$ is scored by the accuracy / negative-error
reward of the personas it induces on each meta-train user's own held-out query
slice (\cref{eq:peruser,eq:meta}). The best $\pi$ on the disjoint meta-validation
users is frozen as $\pi^{\star}$ and applied zero-shot to each held-out test
user: one frozen-LLM call distills the user's support into a persona
$\theta_u$, and one call answers the query under that persona. The backbone is
frozen throughout; only the textual prompt changes.}
\label{fig:arch}
\end{figure*}
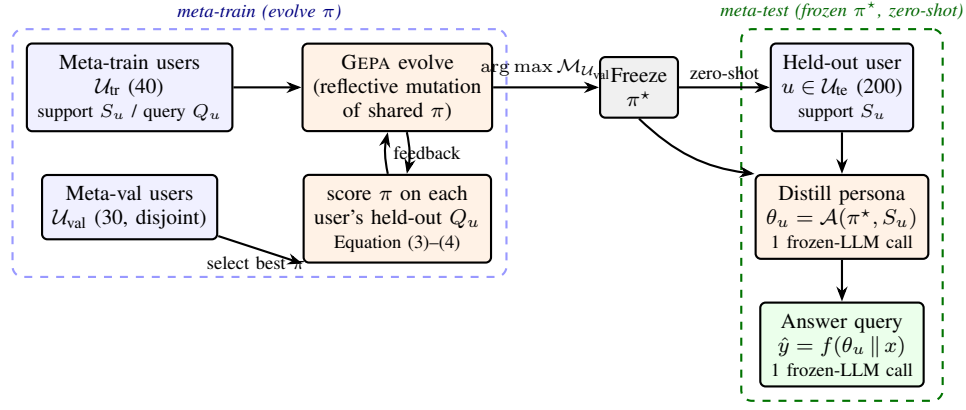

\textsc{Muse} (Meta-learned User-adaptation via Shared Evolution) is the most
direct instantiation of the formulation in \cref{sec:problem} that is compatible
with a hosted, frozen backbone. It meta-learns a \emph{single} shared
adaptation prompt over a user population, freezes it, and applies it zero-shot to
unseen users. The only quantity that differs between \textsc{Muse} and a
fixed-instruction persona baseline such as PAG is \emph{where the adaptation
instruction comes from}: PAG uses a hand-written instruction; \textsc{Muse} uses
a meta-learned one. \Cref{fig:arch} sketches the pipeline.

\subsection{Inner Loop: Persona Distillation under $\pi$}
Given a user's support set, the inner loop issues a single call to the frozen
model in which $\pi$ is injected as the controlling instruction over a digest of
the user's labeled history, and the model returns a bounded persona in a fixed
JSON schema with three fields: a $\le\!600$-character preference summary, up to
six induced exemplars (each an input-gist $\to$ label pair), and up to six short
decision rules of the form ``if \emph{cue} then \emph{label}''. The persona is
clipped to these bounds, rendered into a compact natural-language preamble, and
reused for every query of that user. The schema, the parser, the clipping, and
the answer call are shared verbatim with the PersonaLink/PAG harness so that the
\emph{only} delta across these methods is the adaptation instruction. The
task-specific information (the $15$ \Lampii{} categories or the $1$--$5$ rating
scale) is supplied by the data digest and the fixed template, not by $\pi$, so
that a single $\pi$ can in principle serve both benchmarks. Answering is a single
greedy call with the persona prepended to the query; benchmark scoring is
deterministic, with no LLM judge.

\subsection{Outer Loop: Shared Prompt Evolution}
The meta-parameter $\pi$ is optimized with \textsc{Gepa}'s genetic--Pareto
reflective evolution~\cite{agrawal2025gepa}, reused without modification to its
core loop. The optimizer maintains a growing pool of candidate prompts, seeded
with the hand-written $\pizero$. Each iteration samples a parent from the pool by
instance-frontier (Pareto) coverage over the meta-validation users, draws a
meta-train minibatch, and---if the parent is not already perfect on the
minibatch---reflectively mutates it: the same frozen model is shown the current
$\pi$ together with per-user digests of where the $\pi$-built personas
mispredicted, and is asked to rewrite $\pi$ into a better \emph{general} recipe
for persona construction (how to infer stable preferences from limited history,
surface label-specific cues and user biases, turn patterns into crisp rules and
well-chosen exemplars, and break near-ties), under an explicit instruction not to
specialize to any single user. A child is accepted if it does not worsen the
minibatch score; accepted children are re-scored on the full meta-validation set
and added to the pool. After the rollout budget is exhausted, the candidate with
the best meta-validation aggregate is frozen as $\pistar$.
\Cref{alg:muse} states the procedure. Two design points are essential to the
honesty of the test. First, the reflection model, the inner model, and the answer
model are the \emph{same} frozen backbone, so no extra capacity is smuggled in.
Second, $\pi$ is \emph{shared} across all users and then frozen; it is never tuned
to a test user, which is what makes \textsc{Muse} a meta-learner rather than a
per-user optimizer.

\begin{algorithm}[t]
\caption{\textsc{Muse}: shared-prompt meta-learning for frozen-LLM
personalization.}
\label{alg:muse}
\begin{algorithmic}[1]
\REQUIRE frozen LLM $f$; meta-train users $\mathcal{U}_{\text{tr}}$, meta-val
 $\mathcal{U}_{\text{val}}$; seed prompt $\pizero$; support cap $K$; query size;
 rollout budget $B$
\STATE pool $\leftarrow \{\pizero\}$;\; record $\mathcal{M}_{\mathcal{U}_{\text{val}}}(\pizero)$ via \eqref{eq:meta}
\WHILE{calls $< B$}
  \STATE $\pi \leftarrow$ \textsc{ParetoSelect}(pool, $\mathcal{U}_{\text{val}}$)
    \hfill // instance-frontier coverage
  \STATE draw meta-train minibatch $M\subseteq\mathcal{U}_{\text{tr}}$
  \FORALL{$u \in M$}
    \STATE split profile into support $S_u$ (cap $K$) and held-out query $Q_u$
    \STATE $\theta_u \leftarrow \mathcal{A}(\pi, S_u)$ \hfill // \eqref{eq:inner}: 1 call
    \STATE score $\theta_u$ on $Q_u$; collect mispredictions \hfill // \eqref{eq:peruser}
  \ENDFOR
  \IF{minibatch score is not perfect}
    \STATE $\pi' \leftarrow$ \textsc{ReflectMutate}$(f,\pi,\{\text{mispredictions}\})$
    \STATE accept $\pi'$ if it does not worsen the minibatch score
    \STATE if accepted: score $\pi'$ on $\mathcal{U}_{\text{val}}$; add to pool
  \ENDIF
\ENDWHILE
\STATE $\pistar \leftarrow \arg\max_{\pi\in\text{pool}} \mathcal{M}_{\mathcal{U}_{\text{val}}}(\pi)$
\STATE \textbf{freeze} $\pistar$
\STATE \textbf{at test:} for each $u\in\mathcal{U}_{\text{te}}$, build
  $\theta_u=\mathcal{A}(\pistar,S_u)$ and answer $\hat{y}=f(\theta_u\Vert x)$
\RETURN $\pistar$, meta-test predictions
\end{algorithmic}
\end{algorithm}

\subsection{Controls Built into the Method}
Two controlled variants of \textsc{Muse} are central to the diagnosis and are run
through the identical pipeline. The \textbf{seed} control (\textsc{Muse-seed})
skips evolution entirely and applies the frozen, un-meta-learned $\pizero$ to the
test users; it isolates how much of any apparent gain is generic instruction
quality already present before learning. The \textbf{mismatch} control
(\textsc{Muse-mismatch}) meta-trains $\pi$ under the deranged objective of
\cref{eq:metashuf}---each meta-train user's persona is built from a
deterministically shuffled other user's support set, while still graded on the
user's own queries---then tests on the real users; it isolates whether the
meta-objective carries any cross-user adaptive signal at all. If
\textsc{Muse}$(\pistar)$ cannot beat \textsc{Muse-seed}, the evolution bought
nothing the seed lacked; if it cannot beat \textsc{Muse-mismatch}, the objective
it climbed was structure-blind. Both turn out to hold.

\section{Experimental Setup}\label{sec:setup}

\subsection{Benchmarks and User Splits}
We evaluate on two tasks from the LaMP suite~\cite{salemi2024lamp}.
\textbf{\Lampii{}} is personalized news categorization: given an article, predict
which of $15$ categories \emph{this} user would assign; the metric is accuracy
($\uparrow$) with macro-F1 as a secondary measure. \textbf{\Lampiii{}} is
personalized product-review rating: predict the $1$--$5$ score \emph{this} user
would give; the metrics are mean absolute error (MAE, $\downarrow$) and root mean
squared error (RMSE, $\downarrow$). Each LaMP question identifier corresponds to
one user with a profile of past labeled items. From the development split we
form a candidate user pool by keeping users with at least $16$ profile items and
removing the $200$ designated test users; the meta-train and meta-validation
users are drawn from this pool, yielding three mutually disjoint user sets:
$|\mathcal{U}_{\text{tr}}|{=}40$, $|\mathcal{U}_{\text{val}}|{=}30$, and
$|\mathcal{U}_{\text{te}}|{=}200$ per benchmark. Disjointness is enforced at the
user level, so all reported generalization is across users. Within a user, the
profile is split into a support pool and a held-out query slice that never
overlap; the test users' single LaMP test item is never part of their profile.
\Cref{tab:datastats} summarizes the resulting per-benchmark configuration; the
two tasks share an identical user-split protocol and differ only in label space
(a $15$-way category set versus a $1$--$5$ ordinal scale) and metric family.

\begin{table}[t]
\centering
\caption{Dataset and user-split statistics. Both LaMP tasks share the same
users-as-tasks protocol (disjoint $40$/$30$/$200$ meta-train/val/test users,
$\ge16$-item profile filter); they differ only in the label space and metric
family. ``Support / query'' are the per-user inner-loop slices; the single LaMP
test item per test user is held out of the profile.}
\label{tab:datastats}
\footnotesize
\renewcommand{\arraystretch}{1.18}
\setlength{\tabcolsep}{5pt}
\begin{tabular}{@{}l c c@{}}
\toprule
\textbf{Property} & \textbf{\Lampii{}} & \textbf{\Lampiii{}} \\
\midrule
Task & news categorization & review rating \\
Label space & $15$ categories & $1$--$5$ ordinal \\
Primary metric & accuracy $\uparrow$ & MAE $\downarrow$ \\
Secondary metric & macro-F1 $\uparrow$ & RMSE $\downarrow$ \\
\midrule
Meta-train users $|\mathcal{U}_{\text{tr}}|$ & $40$ & $40$ \\
Meta-val users $|\mathcal{U}_{\text{val}}|$ & $30$ & $30$ \\
Meta-test users $|\mathcal{U}_{\text{te}}|$ & $200$ & $200$ \\
Profile filter & $\ge16$ items & $\ge16$ items \\
Support cap $K$ (default) & $24$ & $24$ \\
Held-out query / user & $8$ & $8$ \\
Test items / test user & $1$ & $1$ \\
User-level disjointness & yes & yes \\
\bottomrule
\end{tabular}
\end{table}

\subsection{Backbone and Decoding}
All methods share one frozen backbone, Qwen3-30B-A3B~\cite{yang2025qwen3}, served
locally on four GPUs. Decoding is greedy (temperature $0$, single sample) for
every call---persona distillation, reflection, and answering---so that the only
per-method difference is the text prepended to an identical question. We use no
LLM judge; all scoring is the benchmark's deterministic metric. Because the
serving stack is not bitwise-deterministic across runs, we treat the
\emph{item} as the unit of analysis (rather than averaging over decoding seeds)
and report paired, per-item statistics, following our harness's convention.

\subsection{Baselines}
We compare \textsc{Muse} against a ladder of training-free and
prompt-optimization baselines, all in the same harness.
\textbf{NoPers} answers with no personalization (a zero-knowledge lower bound).
\textbf{RandomUser} builds the persona from a \emph{random other} user's profile,
controlling for the mere presence of persona text.
\textbf{PAG}~\cite{richardson2023pag} distills the user's profile into a
natural-language persona with a fixed hand-written instruction.
\textbf{PersonaLink} is a self-refining persona distiller; we report its first
refinement round (\textsc{PersonaLink}$_{r1}$), with additional rounds in the
appendix.
\textbf{RAG-$k$} retrieves the user's $k$ most relevant past items and places
them verbatim in the context as few-shot exemplars, for
$k\in\{1,3,5,10,20\}$; this is the strongest training-free baseline and the one
that preserves fine-grained instance-level signal.
The \textbf{Oracle} row in \cref{sec:analysis} is a per-item best-of-three over
$\{\pistar, \text{seed}, \text{mismatch}\}$ and is reported only as an upper
bound on phrasing variance, not as a deployable method.

\subsection{Metrics and Statistics}
Point metrics are accuracy/macro-F1 (\Lampii{}) and MAE/RMSE (\Lampiii{}) over
the $200$ test users. We attach a $95\%$ bootstrap confidence interval to every
metric. Paired comparisons use the item as the unit: for classification we use
McNemar's test on per-item correctness (reporting the discordant counts
$A_{\text{w}}/B_{\text{w}}$), and for regression a paired bootstrap ($10{,}000$
resamples) on per-item absolute error, with the sign of $\Delta$MAE indicating
which method is better. We label any comparison with $p\ge 0.05$ as \ns{}
(not significant). Rank correlations between meta-validation and meta-test use
Spearman $\rho$ and Kendall $\tau$. For the collapse analysis we recompute the
meta-objective \eqref{eq:meta} and the deranged objective \eqref{eq:metashuf} on
the $30$ meta-validation users with a paired bootstrap on their difference.

\subsection{Seed and Mismatch Controls}
The two controls of \cref{sec:method} are run end-to-end. The seed control
applies the frozen $\pizero$ with no evolution. The mismatch control evolves
$\pi$ under a fixed, deterministic derangement of the meta-train user-to-support
assignment (no fixed points), tests on the real users, and is otherwise
identical to \textsc{Muse}. We additionally sweep the support cap
$K\in\{4,8,16,32\}$ and the meta-train size $|\mathcal{U}_{\text{tr}}|\in\{10,20,40\}$,
re-evolving and re-testing for each setting, to probe whether more support or
more meta-train users induces transfer.

\section{Results}\label{sec:results}

\Cref{tab:main} reports the main comparison; \cref{fig:teaser} visualizes the
headline rows. We organize the findings as finding-titled subsections.
\Cref{tab:cis} attaches the $95\%$ bootstrap confidence intervals to the two
headline metrics (\Lampii{} accuracy and \Lampiii{} MAE); the intervals are wide
relative to the inter-method spread and overlap heavily across all persona
methods and \textsc{Muse} variants, which is the interval-level restatement of
the null findings below---only \textsc{Rag} on \Lampiii{} sits in a visibly
lower MAE band.

\begin{table}[t]
\centering
\caption{Headline point estimates with $95\%$ bootstrap confidence intervals
($200$ test users; intervals from the same item-level resampling used for the
paired tests). For \Lampii{} the metric is accuracy ($\uparrow$); for \Lampiii{}
it is MAE ($\downarrow$). The persona/\textsc{Muse} intervals mutually overlap
(no learned-adaptation separation), whereas \textsc{Rag-20} on \Lampiii{} clears
the persona band.}
\label{tab:cis}
\footnotesize
\renewcommand{\arraystretch}{1.18}
\setlength{\tabcolsep}{5pt}
\begin{tabular}{@{}l c c@{}}
\toprule
& \textbf{\Lampii{} Acc $\uparrow$} & \textbf{\Lampiii{} MAE $\downarrow$} \\
\textbf{Method} & {[}95\% CI{]} & {[}95\% CI{]} \\
\midrule
NoPers              & $0.645$ \,{\scriptsize$[0.576,0.711]$} & $0.500$ \,{\scriptsize$[0.420,0.585]$} \\
RandomUser          & $0.665$ \,{\scriptsize$[0.597,0.730]$} & $0.565$ \,{\scriptsize$[0.475,0.660]$} \\
PAG                 & $0.745$ \,{\scriptsize$[0.685,0.805]$} & $0.450$ \,{\scriptsize$[0.370,0.535]$} \\
PersonaLink$_{r1}$  & $0.755$ \,{\scriptsize$[0.695,0.810]$} & $0.425$ \,{\scriptsize$[0.350,0.505]$} \\
RAG-5               & $0.765$ \,{\scriptsize$[0.705,0.820]$} & $0.285$ \,{\scriptsize$[0.220,0.355]$} \\
RAG-20              & $\mathbf{0.790}$ \,{\scriptsize$[0.735,0.845]$} & $\mathbf{0.250}$ \,{\scriptsize$[0.190,0.315]$} \\
\midrule
\textsc{Muse} ($\pistar$)      & $0.765$ \,{\scriptsize$[0.705,0.820]$} & $0.425$ \,{\scriptsize$[0.350,0.505]$} \\
\textsc{Muse-seed} ($\pizero$) & $0.775$ \,{\scriptsize$[0.715,0.830]$} & $0.400$ \,{\scriptsize$[0.325,0.480]$} \\
\textsc{Muse-mismatch}         & $0.740$ \,{\scriptsize$[0.680,0.800]$} & $0.425$ \,{\scriptsize$[0.350,0.505]$} \\
\bottomrule
\end{tabular}
\end{table}

\begin{table*}[t]
\centering
\caption{Main results over $200$ held-out users per benchmark. \Lampii{}:
accuracy and macro-F1 ($\uparrow$). \Lampiii{}: MAE and RMSE ($\downarrow$).
\textsc{Muse} variants are grouped at the bottom. \textbf{Bold} marks the best
value in each column among deployable methods (the per-item Oracle is an upper
bound on phrasing variance, not a method). Key paired tests
(\cref{sec:results,sec:analysis}): \textsc{Muse} vs.\ \textsc{Muse-seed} and vs.\
\textsc{Muse-mismatch} are \emph{not significant} on both benchmarks;
\textsc{Muse} vs.\ \textsc{Rag-20} is \ns{} on \Lampii{} but
$\Delta\text{MAE}{=}{+}0.175$, $95\%$\,CI $[{+}0.095,{+}0.260]$, $p<0.001$ on
\Lampiii{} (retrieval wins regression). \textbf{Cost} is \#LLM calls per query.
Significance vs.\ \textsc{Muse}($\pi^\star$), per-item paired tests (McNemar for
accuracy, paired bootstrap $10^4$ for MAE): $^{***}p<0.001$, $^{**}p<0.01$,
$^{*}p<0.05$, $^{\textsc{ns}}$ not significant ($p\ge0.05$), $^{=}$ exact tie.}
\label{tab:main}
\footnotesize
\renewcommand{\arraystretch}{1.12}
\setlength{\tabcolsep}{6pt}
\begin{tabular}{@{}l l c cc cc@{}}
\toprule
& & & \multicolumn{2}{c}{\textbf{\Lampii{} (categorization)}}
  & \multicolumn{2}{c}{\textbf{\Lampiii{} (rating)}} \\
\cmidrule(lr){4-5}\cmidrule(lr){6-7}
\textbf{Method} & \textbf{Type} & \textbf{Cost}
  & Acc $\uparrow$ & macro-F1 $\uparrow$
  & MAE $\downarrow$ & RMSE $\downarrow$ \\
\midrule
NoPers              & none      & 1 & 0.645 & 0.388 & 0.500 & 0.794 \\
RandomUser          & control   & 1 & 0.665 & 0.393 & 0.565 & 0.908 \\
\midrule
PAG                 & persona   & 2 & 0.745$^{\textsc{ns}}$ & 0.485 & 0.450 & 0.742 \\
PersonaLink$_{r1}$  & persona   & 2 & 0.755$^{\textsc{ns}}$ & 0.494 & 0.425 & 0.731 \\
\midrule
RAG-1               & retrieval & 1 & 0.695 & 0.531 & 0.415 & 0.711 \\
RAG-3               & retrieval & 1 & 0.760 & 0.607 & 0.290 & 0.616 \\
RAG-5               & retrieval & 1 & 0.765 & \textbf{0.608} & 0.285 & 0.621 \\
RAG-10              & retrieval & 1 & 0.785 & 0.601 & 0.290 & 0.608 \\
RAG-20              & retrieval & 1 & \textbf{0.790}$^{\textsc{ns}}$ & 0.550 & \textbf{0.250}$^{***}$ & \textbf{0.566}$^{***}$ \\
\midrule
\textbf{\textsc{Muse}} ($\pistar$) & meta-learn & 2 & 0.765 & 0.501 & 0.425 & 0.731 \\
\textsc{Muse-seed} ($\pizero$)     & seed ctrl  & 2 & 0.775$^{\textsc{ns}}$ & 0.449 & 0.400 & 0.731 \\
\textsc{Muse-mismatch}             & derange ctrl & 2 & 0.740$^{\textsc{ns}}$ & 0.433 & 0.425$^{=}$ & 0.731 \\
\midrule
\textit{Oracle}$_{\{\pistar,\text{seed},\text{mis}\}}$ & \textit{u.b.} & \textit{--} & \textit{0.860} & \textit{0.526} & \textit{0.245} & \textit{--} \\
\bottomrule
\end{tabular}
\end{table*}

\subsection{Finding 1: Meta-Learning Does Not Beat Its Own Seed}
The central comparison is \textsc{Muse}$(\pistar)$ against \textsc{Muse-seed},
which differ \emph{only} in whether the adaptation prompt was evolved. Evolution
does not help. On \Lampii{}, \textsc{Muse} reaches accuracy $0.765$ versus the
seed's $0.775$; the McNemar test on per-item correctness gives discordant counts
$13/15$ in the seed's favor with $p{=}0.850$ (\ns{}). On \Lampiii{}, \textsc{Muse}
attains MAE $0.425$ versus the seed's $0.400$, a paired-bootstrap
$\Delta\text{MAE}{=}{+}0.025$, $95\%$\,CI $[-0.045,+0.095]$, $p{=}0.523$ (\ns{}).
In both cases the un-evolved seed is numerically \emph{better}, and the
difference is statistically indistinguishable from zero. Whatever \textsc{Gepa}
extracted over the user population, it is not an improvement in per-user
adaptation that survives transfer to new users.

\subsection{Finding 2: Meta-Learning Does Not Beat a Structure-Broken Control}
\textsc{Muse-mismatch} meta-trains the shared prompt on \emph{deliberately wrong}
user--support pairs, so any genuine cross-user signal in the objective has been
destroyed by construction; if \textsc{Muse} learned real adaptive structure it
should beat this control. It does not. On \Lampii{}, \textsc{Muse} $0.765$ versus
mismatch $0.740$ is \ns{} (McNemar $16/11$, $p{=}0.441$); on \Lampiii{} the two
\emph{tie} at MAE $0.425$ ($\Delta\text{MAE}{=}0.000$, CI $[-0.075,+0.080]$,
$p{=}1.000$). Meta-training under a scrambled correspondence is as good as
meta-training under the real one---a direct symptom of the collapse we dissect in
\cref{sec:analysis}.

\subsection{Finding 3: Parity With Persona Baselines on Categorization}
On \Lampii{}, every distilled-persona method---\textsc{Muse}, PAG, and
PersonaLink---clusters in a narrow band around $0.74$--$0.78$ accuracy and is
mutually indistinguishable: \textsc{Muse} vs.\ PersonaLink$_{r1}$ ($0.765$ vs.\
$0.755$) is \ns{} (McNemar $17/15$, $p{=}0.860$), and \textsc{Muse} vs.\ PAG
($0.765$ vs.\ $0.745$) is \ns{} ($18/14$, $p{=}0.596$). The meta-learned prompt
neither helps nor hurts relative to a hand-written one. All persona methods do
clear the un-personalized and random-user floors (NoPers $0.645$, RandomUser
$0.665$), so the personas \emph{are} doing something; that ``something'' is simply
not improved by meta-learning the instruction.

\subsection{Finding 4: Retrieval Dominates on Regression}
\begin{figure*}[t]
\centering
\IfFileExists{figures/fig_ragscaling.pdf}{%
  \includegraphics[width=0.92\textwidth]{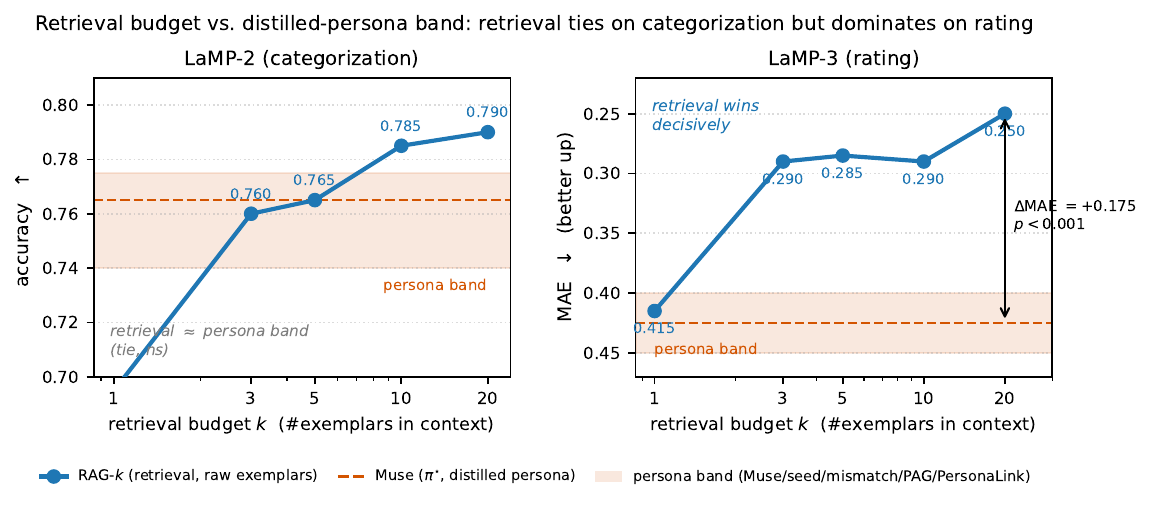}}{%
  \fbox{\parbox[c][4cm][c]{0.9\textwidth}{\centering \texttt{fig\_ragscaling.pdf}}}}
\caption{Retrieval budget versus the distilled-persona band (numbers from
\cref{tab:main}). \emph{Left:} \Lampii{} accuracy ($\uparrow$); the
retrieval-augmented few-shot curve RAG-$k$ rises with $k$ but stays inside the
shaded persona band (\textsc{Muse}/seed/mismatch/PAG/PersonaLink span
$0.740$--$0.775$), so retrieval and personas are statistically tied (RAG-20
$0.790$ vs.\ \textsc{Muse} $0.765$, \ns{}). \emph{Right:} \Lampiii{} MAE
($\downarrow$, plotted so better is up); even RAG-3/5 ($\text{MAE}\approx0.29$)
leave the persona band, and RAG-20 ($0.250$) opens a significant
$\Delta\text{MAE}{=}{+}0.175$ gap over \textsc{Muse} ($0.425$, $p<0.001$).
Distilling a user into a short persona discards the instance-level rating
calibration that raw exemplars preserve; the effect is large only on the ordinal
task.}
\label{fig:ragscaling}
\end{figure*}

The one large, significant effect in the entire study runs \emph{against} the
persona paradigm. On \Lampiii{}, retrieval-augmented few-shot prompts dominate:
RAG-20 attains MAE $0.250$ (RMSE $0.566$) versus \textsc{Muse}'s $0.425$, a
paired $\Delta\text{MAE}{=}{+}0.175$, $95\%$\,CI $[{+}0.095,{+}0.260]$,
$p<0.001$. Even modest retrieval budgets (RAG-3/5 at MAE $\approx\!0.29$) beat all
persona methods decisively. The asymmetry with \Lampii{}, where retrieval is only
marginally and non-significantly ahead (RAG-20 $0.790$ vs.\ \textsc{Muse}
$0.765$, \ns{}), is informative: distilling a user into a short natural-language
persona discards the fine-grained rating calibration that raw few-shot exemplars
preserve. We report this prominently rather than minimize it; on rating, the
representation that keeps instance-level signal wins, independently of whether the
adaptation policy is learned.
\Cref{fig:ragscaling} plots the retrieval ladder against the distilled-persona
band on both tasks and makes the asymmetry visual: on \Lampii{} the RAG curve
threads straight through the persona band (a tie), whereas on \Lampiii{} it
pulls clear of the band for every $k\ge3$ and reaches its widest, significant gap
at $k{=}20$. \Cref{tab:sigtests} consolidates the head-to-head paired tests that
back the four findings---each $\Delta$, its $95\%$ confidence interval, and its
$p$-value---in one place, so the reader can verify at a glance that every
persona-vs-\textsc{Muse} comparison is \ns{} while only \textsc{Rag-20} on
\Lampiii{} clears significance.

\begin{table*}[t]
\centering
\caption{Consolidated paired significance tests behind
\cref{sec:results} (Findings 1--4). Each row is a comparison against the listed
reference; $\Delta$ is (reference $-$ \textsc{Muse}) for accuracy and
(\textsc{Muse} $-$ reference) for MAE, signed so a positive value favors
\textsc{Muse}'s competitor. Classification rows report McNemar discordant counts
$A_{\text{w}}/B_{\text{w}}$ on per-item correctness; regression rows report a
paired bootstrap ($10^4$ resamples) on per-item absolute error. The only
comparison that clears $p<0.05$ is \textsc{Rag-20} on \Lampiii{}; every
\textsc{Muse}-vs-persona and \textsc{Muse}-vs-control comparison is \ns{}.}
\label{tab:sigtests}
\footnotesize
\renewcommand{\arraystretch}{1.18}
\setlength{\tabcolsep}{6pt}
\begin{tabular}{@{}l l c c c c c@{}}
\toprule
\textbf{Bench} & \textbf{Comparison (vs.\ \textsc{Muse})} & \textbf{$\Delta$}
  & \textbf{95\% CI} & \textbf{$A_{\text{w}}/B_{\text{w}}$} & \textbf{$p$}
  & \textbf{Verdict} \\
\midrule
\Lampii{}  & \textsc{Muse-seed} ($\pizero$)      & $+0.010$ & $[-0.045,+0.065]$ & $13/15$ & $0.850$ & \ns{} \\
\Lampii{}  & \textsc{Muse-mismatch}              & $-0.025$ & $[-0.085,+0.040]$ & $16/11$ & $0.441$ & \ns{} \\
\Lampii{}  & PAG                                 & $-0.020$ & $[-0.080,+0.045]$ & $18/14$ & $0.596$ & \ns{} \\
\Lampii{}  & PersonaLink$_{r1}$                  & $-0.010$ & $[-0.070,+0.050]$ & $17/15$ & $0.860$ & \ns{} \\
\Lampii{}  & \textsc{Rag-20}                     & $+0.025$ & $[-0.035,+0.085]$ & $21/16$ & $0.515$ & \ns{} \\
\midrule
\Lampiii{} & \textsc{Muse-seed} ($\pizero$)      & $+0.025$ & $[-0.045,+0.095]$ & \textemdash & $0.523$ & \ns{} \\
\Lampiii{} & \textsc{Muse-mismatch}              & $+0.000$ & $[-0.075,+0.080]$ & \textemdash & $1.000$ & \ns{} \\
\Lampiii{} & PAG                                 & $-0.025$ & $[-0.100,+0.050]$ & \textemdash & $0.520$ & \ns{} \\
\Lampiii{} & PersonaLink$_{r1}$                  & $+0.000$ & $[-0.078,+0.078]$ & \textemdash & $1.000$ & \ns{} \\
\Lampiii{} & \textsc{Rag-20}                     & $+0.175$ & $[+0.095,+0.260]$ & \textemdash & $<0.001$ & $^{***}$ \\
\bottomrule
\end{tabular}
\end{table*}

\section{Analysis: Meta-Objective Collapse}\label{sec:analysis}

The results of \cref{sec:results} are not four separate nulls; they are
consequences of one mechanism. We establish \emph{meta-objective collapse}
through three mechanistic legs---(i)~no transfer over the seed, (ii)~a
structure-blind objective, and (iii)~meta-overfitting at small population
scale---and then decompose the residual test behavior with an invariance/oracle
analysis and a semantic diff of the evolved prompts.

\subsection{Leg I: No Transfer --- $\pistar$ Behaves Like the Seed}
If evolution had learned transferable structure, $\pistar$ would behave
differently---and better---than $\pizero$ on held-out users. Instead the two are
behaviorally near-identical. \Cref{fig:invariance} (left) decomposes the $200$
test users by how the prediction moves across $\{\pistar,\text{seed},
\text{mismatch}\}$. On \Lampii{}, $80.0\%$ of users are \emph{phrasing-invariant}
($132$ invariant-correct, $28$ invariant-wrong) and only $40$ ($20\%$) ever
``swing''; on \Lampiii{}, $72.0\%$ are invariant ($97$/$47$) with $56$ ($28\%$)
swinging. Among the swing users---the only ones the choice of prompt can affect---
$\pistar$ and the seed split essentially evenly: McNemar $13/15$, $p{=}0.85$ on
\Lampii{} and $13/19$, $p{=}0.38$ on \Lampiii{}, both \ns{}, with the seed
numerically ahead. The evolved prompt is not a different adaptation policy; it is
a paraphrase of the seed that the frozen backbone treats almost identically.

\begin{figure*}[t]
\centering
\IfFileExists{figures/fig_collapse.pdf}{%
  \includegraphics[width=0.86\textwidth]{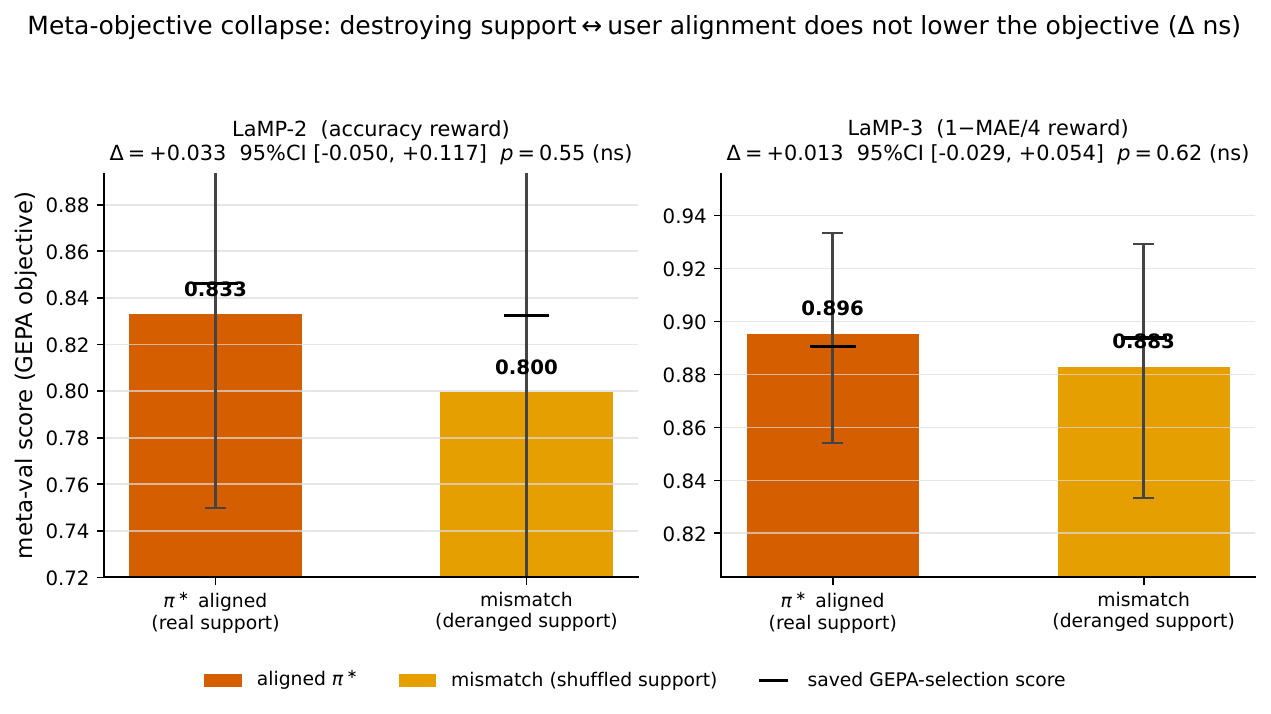}}{%
  \fbox{\parbox[c][4cm][c]{0.8\textwidth}{\centering \texttt{fig\_collapse.pdf}}}}
\caption{\textbf{The money figure: the meta-objective is structure-blind.}
Recomputed meta-validation objective on the $30$ meta-val users for the evolved
prompt with \emph{aligned} (real) supports versus a \emph{mismatched} (deranged)
support assignment, \cref{eq:meta} vs.\ \eqref{eq:metashuf}. Destroying the
user--support correspondence does \emph{not} lower the objective: the aligned
score is (numerically) at least the mismatch score on \emph{both} benchmarks and
neither gap is significant---$0.833$ aligned vs.\ $0.800$ mismatch,
$\Delta{=}{+}0.033$, $95\%$\,CI $[-0.050,{+}0.117]$, $p{=}0.55$ (\ns{}) on
\Lampii{}, and $0.896$ vs.\ $0.883$, $\Delta{=}{+}0.013$, $[-0.029,{+}0.054]$,
$p{=}0.62$ (\ns{}) on \Lampiii{} (the saved selection scores agree:
$0.846$/$0.832$ and $0.891$/$0.894$). An objective statistically invariant to
whether the correspondence is real cannot be optimized into transferable
adaptation.}
\label{fig:collapse}
\end{figure*}

\subsection{Leg II: A Structure-Blind Objective (the Money Figure)}
The decisive evidence is in the objective itself. We recompute, on the $30$
meta-validation users, the aligned meta-objective \eqref{eq:meta} and its
deranged counterpart \eqref{eq:metashuf} for the evolved prompt, and compare them
with a paired bootstrap on the per-user difference (\cref{fig:collapse}). The two
are statistically indistinguishable, and on \emph{both} benchmarks the aligned
objective is (numerically) at least as high as the mismatched one. On \Lampii{}
the aligned score is $0.833$ versus mismatch $0.800$, a difference of
$\Delta{=}{+}0.033$, $95\%$\,CI $[-0.050,+0.117]$, $p{=}0.555$ (\ns{}); on
\Lampiii{} the aligned score is $0.896$ versus mismatch $0.883$,
$\Delta{=}{+}0.013$, $[-0.029,+0.054]$, $p{=}0.622$ (\ns{}). The \textsc{Gepa}
selection-time scores tell the same story (the saved selection scores agree:
$0.846$/$0.832$ on \Lampii{} and $0.891$/$0.894$ on \Lampiii{}): aligned
$\approx$ mismatch, and the objective is blind to the correspondence. This is
meta-objective collapse in the precise sense of
\cref{eq:collapse}: the quantity the optimizer climbs does not distinguish a
user's persona built from their own history from one built from a stranger's. An
optimizer maximizing such a quantity cannot, even in principle, acquire
cross-user adaptive structure---it can only move along directions to which the
objective is sensitive, namely generic phrasing. This is the operational meaning
of \emph{proxy-objective misalignment}: the optimized proxy and the intended goal
(transferable adaptation) have come apart, and the derangement control measures
exactly the size of the gap (here, indistinguishable from zero).

\subsection{Leg III: Meta-Overfitting at Small Population Scale}
\begin{figure}[t]
\centering
\IfFileExists{figures/fig_gengap.pdf}{%
  \includegraphics[width=\columnwidth]{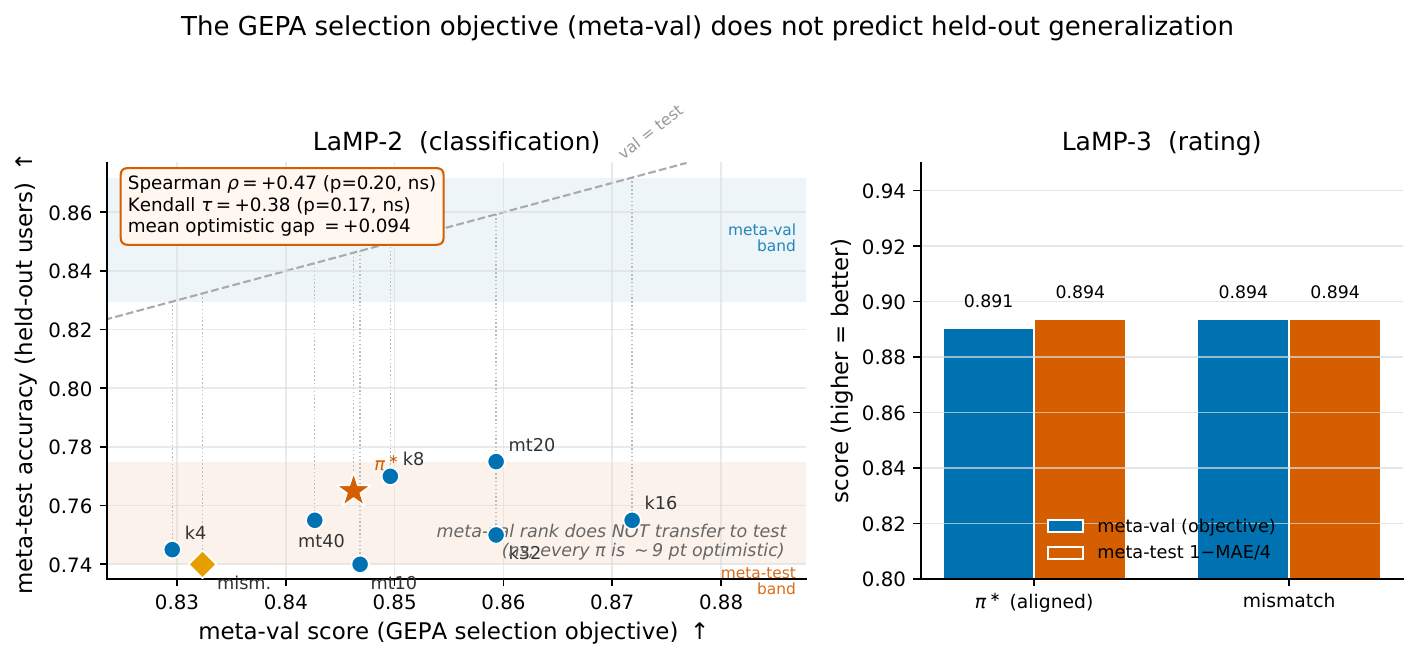}}{%
  \fbox{\parbox[c][3.4cm][c]{0.92\columnwidth}{\centering \texttt{fig\_gengap.pdf}}}}
\caption{Meta-overfitting. \emph{Left:} meta-validation versus meta-test accuracy
for nine \Lampii{} variants; every point sits below $y{=}x$ (mean optimistic gap
$+0.094$), the rank correlation is positive but not significant
($\rho{=}+0.47$, $p{=}0.20$; $\tau{=}+0.38$, $p{=}0.17$), and the highest
meta-val point ($K{=}16$) has below-median test accuracy. \emph{Right:}
\Lampiii{} meta-val objective versus the held-out $1{-}\text{MAE}/4$ reward for
$\pistar$ and the mismatch control are flat and near-equal. The selection
objective does not predict held-out generalization.}
\label{fig:gengap}
\end{figure}

The little that does separate candidate prompts is not predictive of held-out
performance---it is selection-set overfitting. \Cref{fig:gengap} plots
meta-validation against meta-test for nine \Lampii{} variants (the seed, the
mismatch control, and the $K$- and meta-train-size sweeps). Three facts stand
out. First, there is a uniform \emph{optimistic gap}: meta-test trails
meta-validation by $\approx\!+0.094$ on average (every variant lies below the
$y{=}x$ line), exactly the over-optimism expected when a small validation
population is used to select among prompts. Second, meta-validation does
\emph{not} significantly rank meta-test: Spearman $\rho{=}+0.47$ ($p{=}0.20$) and
Kendall $\tau{=}+0.38$ ($p{=}0.17$), both \ns{} and positive (we explicitly do
\emph{not} claim a negative correlation); strikingly, the best meta-validation
variant ($K{=}16$ at $0.872$) yields a \emph{below-median} test accuracy of
$0.755$. Third, the sweeps are flat and non-monotone: accuracy wobbles inside
$0.745$/$0.770$/$0.755$/$0.750$ across $K\in\{4,8,16,32\}$, and
$0.740$/$0.775$/$0.755$ across $|\mathcal{U}_{\text{tr}}|\in\{10,20,40\}$, with no
trend that survives the $\pm0.06$ McNemar noise band. More support and more
meta-train users do not buy transfer; there is no scaling signal to ride.

\subsection{Invariance and Oracle Decomposition}
\begin{figure}[t]
\centering
\IfFileExists{figures/fig_invariance.pdf}{%
  \includegraphics[width=\columnwidth]{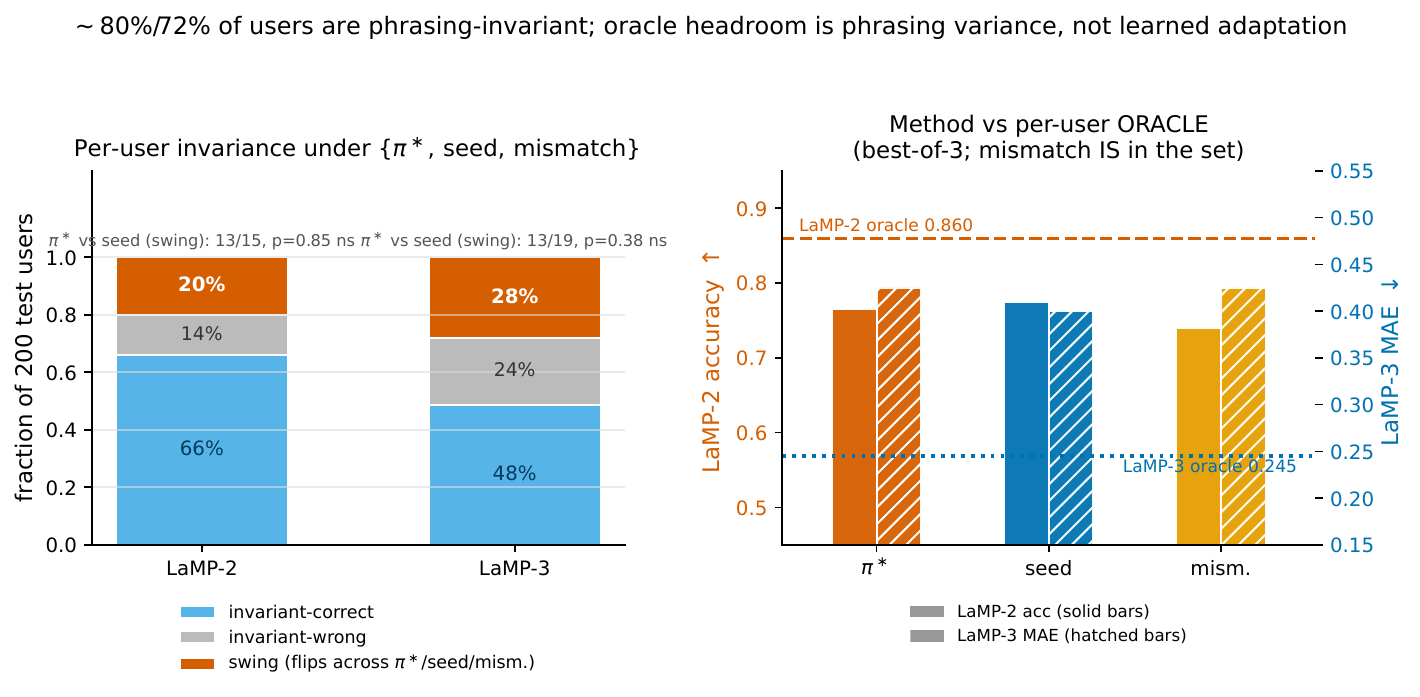}}{%
  \fbox{\parbox[c][3.4cm][c]{0.92\columnwidth}{\centering \texttt{fig\_invariance.pdf}}}}
\caption{Invariance and oracle. \emph{Left:} fraction of the $200$ test users that
are invariant-correct, invariant-wrong, or swing across
$\{\pistar,\text{seed},\text{mismatch}\}$; $80\%$/$72\%$ are phrasing-invariant on
\Lampii{}/\Lampiii{}, and among swing users $\pistar$ does not beat the seed
(McNemar $13/15$, $p{=}0.85$; $13/19$, $p{=}0.38$; both \ns{}). \emph{Right:} a
per-item Oracle over the three prompts reaches $0.860$ accuracy / $0.245$ MAE, but
since the mismatch prompt is in the set, this headroom is phrasing variance, not
learned adaptation.}
\label{fig:invariance}
\end{figure}

\Cref{fig:invariance} (right) bounds what \emph{any} choice among
$\{\pistar,\text{seed},\text{mismatch}\}$ could buy. A per-item Oracle that picks
the best of the three reaches accuracy $0.860$ on \Lampii{} and MAE $0.245$ on
\Lampiii{}, well above each individual method. It is tempting to read this
headroom as latent adaptation waiting to be unlocked. It is not. The
\emph{mismatch} prompt---whose supports were scrambled---is \emph{in} the Oracle
set, so the Oracle is rewarded whenever a scrambled-support persona happens to
phrase the answer in a way the backbone gets right. The headroom is therefore
\emph{phrasing variance} across paraphrases of essentially the same policy, not
learned cross-user adaptation; no selector with access only to a user's own
support could realize it, because the three prompts do not encode different
adaptive behavior, only different wordings. The Oracle bound thus reinforces,
rather than softens, the collapse diagnosis.

\subsection{What the Evolved Prompt Actually Changed}
A semantic analysis of the prompts corroborates the mechanism. Embedding the
prompts with a sentence encoder~\cite{reimers2019sbert}, the evolved $\pistar$ is
only moderately similar to the seed (cosine $0.513$ on \Lampii{}, $0.539$ on
\Lampiii{}), confirming that \textsc{Gepa} did rewrite the text substantially.
But the rewrites do \emph{not} organize by adaptive content. Across all evolved
prompts (the two benchmarks, the mismatch variants, and the sweeps), the prompts
cluster by \emph{benchmark}, not by aligned-versus-mismatch: mean within-benchmark
cosine ($0.723$ on \Lampii{}, $0.729$ on \Lampiii{}) exceeds the cross-benchmark
mean ($0.665$). If the evolution had discovered genuine adaptive structure, the
aligned and mismatched prompts would differ in kind; instead they are as similar
as any two same-benchmark prompts. Reading the diff confirms the impression: the
seed-to-$\pistar$ edit reformats free prose into a structured JSON recipe and adds
generic ``polish'' lines---``focus on stability over short-term fluctuations'',
``select examples that best represent each category''---i.e., better instruction
formatting, not a new adaptation mechanism. The evolution found the directions the
objective rewards (phrasing, formatting) and could not find the direction it does
not reward (cross-user adaptation), exactly as collapse predicts.
\Cref{tab:promptgeo} lays out the cosine geometry that grounds this reading: the
seed-to-$\pistar$ similarity is only moderate (the rewrite is real), yet the
aligned and mismatch prompts sit at within-benchmark similarity, and the
within-benchmark similarity exceeds the cross-benchmark similarity on both tasks.
The prompts organize by \emph{benchmark}, not by whether their training supports
were aligned---the signature of a benchmark-shaped phrasing edit rather than a
discovered adaptive mechanism.

\begin{table}[t]
\centering
\caption{Semantic geometry of the evolved prompts (cosine similarity of
sentence-encoder embeddings~\cite{reimers2019sbert}). The seed-to-$\pistar$
similarity is only moderate (substantial rewrite), but the aligned-vs-mismatch
similarity is at the within-benchmark level, and within-benchmark exceeds
cross-benchmark. Prompts cluster by benchmark, not by whether their meta-training
supports were aligned---consistent with a phrasing/formatting edit, not a learned
adaptation mechanism.}
\label{tab:promptgeo}
\footnotesize
\renewcommand{\arraystretch}{1.18}
\setlength{\tabcolsep}{5pt}
\begin{tabular}{@{}l c c@{}}
\toprule
\textbf{Cosine comparison} & \textbf{\Lampii{}} & \textbf{\Lampiii{}} \\
\midrule
seed $\pizero$ vs.\ evolved $\pistar$      & $0.513$ & $0.539$ \\
aligned $\pistar$ vs.\ mismatch prompt      & $0.731$ & $0.726$ \\
mean within-benchmark (all variants)        & $0.723$ & $0.729$ \\
\midrule
mean cross-benchmark (all variants)         & \multicolumn{2}{c}{$0.665$} \\
\bottomrule
\end{tabular}
\end{table}

\section{Discussion}\label{sec:discussion}

\subsection{A Reusable Evaluation Protocol}
The most transferable product of this study is methodological. The same three
controls that diagnose collapse here can pre-empt false positives anywhere
prompt-space methods claim to ``learn to adapt'' to users or tasks.
\textbf{(1)~Seed control:} run the identical pipeline with the un-optimized seed
prompt; if the optimized prompt does not beat it, the optimization bought
instruction polish, not learning. \textbf{(2)~Wrong-support (derangement)
control:} recompute the selection objective with the support-to-user
correspondence scrambled; if the objective does not drop, it is structure-blind
and cannot encode adaptation. \textbf{(3)~Invariance/oracle decomposition:}
measure how many test instances are invariant to the choice of prompt, and check
whether any oracle headroom comes from genuinely different policies or merely from
paraphrase variance (which is exposed by including a structure-broken variant in
the oracle set). Each control is cheap---it reuses the existing pipeline---and
each isolates a distinct confound (prior instruction quality, objective
structure, phrasing variance). We recommend reporting all three before
attributing gains to learned adaptation. Their absence is, we suspect, why
prompt-space personalization results can look stronger than they are: instruction
polish and small-population validation overfit are easy to mistake for transfer.

\subsection{When Prompt-Space Meta-Learning Can and Cannot Work}
Our negative result is specific, and its boundary is informative. The failure is
not that frozen-LLM personalization is hopeless---the persona methods clearly
beat the un-personalized floor, and retrieval wins outright on rating---but that
\emph{meta-learning a shared adaptation instruction against an aggregate
population objective} does not add transferable structure on top. The diagnosis
points directly at the missing ingredient: \emph{grounded, per-instance feedback}.
Test-time adaptation methods that succeed do so because their objective is tied to
a checkable signal on each instance---a verifier, a self-consistency vote, a tool
execution. Grounded agentic adaptation, where each step is verified against
per-call execution feedback (\cref{sec:related}), improves precisely because its
reward is instance-level and structure-bearing. \textsc{Muse}'s meta-objective is the
opposite: a population average with no per-instance grounding, which is why
scrambling the correspondence leaves it unchanged. We therefore expect
prompt-space meta-learning to transfer when, and roughly only when, the
meta-objective can be made sensitive to the user--instance correspondence---for
example by per-user verifiable rewards, contrastive objectives that explicitly
penalize wrong-user personas, or richer support sets that make the right user
distinguishable. Absent such grounding, the objective collapses to phrasing and a
hand-written seed is as good as evolution. This also reconciles our finding with
the implicit-meta-learning view of ICL: the frozen backbone already adapts to a
user's in-context exemplars, so an explicit, ungrounded prompt-space objective has
little additional structure to grab.

\subsection{Threats to Validity and Limitations}
We delimit the claims carefully. \emph{(i)~Backbone.} We use a single backbone
family (Qwen3-30B-A3B); a larger or differently-tuned model might extract signal a
smaller one cannot, though the structure-blindness of the objective
(\cref{eq:collapse}) is a property of the \emph{objective}, not the model, and
would have to be broken by changing the objective, not merely the backbone.
\emph{(ii)~Benchmarks.} We study two LaMP tasks (categorization and rating);
other personalization tasks---generation, dialogue style, long-form
preference---may behave differently, and we do not claim universality.
\emph{(iii)~Prompt class.} \textsc{Muse} meta-learns a \emph{natural-language}
adaptation prompt evolved by \textsc{Gepa}; soft-prompt or gradient-based
meta-learners, or contrastive objectives, are outside our scope and are the
natural next test of whether grounding the objective restores transfer.
\emph{(iv)~Determinism.} The serving stack is not bitwise-deterministic, so we use
the item as the unit of analysis and rely on paired tests; we report \ns{} where
warranted and never convert a non-significant difference into a claim.
\emph{(v)~Scope of the positive claim.} The defensible positive contribution is
the diagnostic protocol and the collapse mechanism, not a new state-of-the-art
personalizer. We present it as such.

\section{Conclusion}\label{sec:conclusion}

We asked whether prompt-space meta-learning genuinely learns to adapt a frozen LLM
across users, and answered, with controls, that on two standard personalization
benchmarks it does not. \textsc{Muse}---a clean instantiation of
\textsc{Gepa}-style shared-prompt evolution as a cross-user meta-learner---fails
to beat its own un-evolved seed prompt, fails to beat a structure-broken control
that meta-trains on scrambled user--support pairs, ties the strongest persona
baselines, and is decisively dominated by plain retrieval on rating. The unifying
cause is \emph{meta-objective collapse}: the aggregate meta-validation objective
is statistically invariant to whether the user--support correspondence is real, so
it carries no exploitable cross-user adaptive signal, and the optimizer can only
buy generic instruction polish (already present in a hand-written seed) plus
overfitting of a small validation population. We quantified each leg---no transfer
over the seed, a structure-blind objective, an $\approx\!9$-point optimistic gap
with non-predictive selection, and $80\%$/$72\%$ phrasing invariance with oracle
headroom attributable to paraphrase variance---and distilled a reusable
evaluation protocol (seed, wrong-support, and invariance/oracle controls) so that
the community can separate learned adaptation from instruction polish and
validation overfit. The constructive reading is a hypothesis with teeth: what
prompt-space meta-learning lacks is grounded, per-instance feedback, and supplying
it is the most promising route to making the meta-objective informative again.

\appendices

\section{Full Sweep Numbers}\label{app:sweeps}
\Cref{tab:sweep} reports the support-size ($K$) and meta-train-size sweeps on
\Lampii{} that underlie \cref{fig:gengap}, including each variant's
\textsc{Gepa}-selection meta-validation score, held-out test accuracy and
macro-F1, and the optimistic gap. Higher meta-validation does not imply higher
test accuracy; the highest meta-val variant ($K{=}16$, $0.872$) has below-median
test accuracy ($0.755$).

\begin{table}[h]
\centering
\caption{\Lampii{} support-size and meta-train-size sweeps ($200$ test users).
``meta-val'' is the \textsc{Gepa}-selection aggregate on the $30$ meta-val users
(the optimized quantity), not the test metric. ``gap'' $=$ meta-val $-$ test
accuracy.}
\label{tab:sweep}
\footnotesize
\renewcommand{\arraystretch}{1.12}
\setlength{\tabcolsep}{5pt}
\begin{tabular}{@{}l ccc ccc@{}}
\toprule
Variant & $K$ & $|\mathcal{U}_{\text{tr}}|$ & meta-val
        & Acc $\uparrow$ & macro-F1 & gap \\
\midrule
\textsc{Muse} ($\pistar$) & 24 & 40 & 0.846 & 0.765 & 0.501 & 0.081 \\
\textsc{Muse-mismatch}    & 24 & 40 & 0.832 & 0.740 & 0.433 & 0.092 \\
\midrule
\textsc{Muse}-$K$4        & 4  & 40 & 0.830 & 0.745 & 0.444 & 0.085 \\
\textsc{Muse}-$K$8        & 8  & 40 & 0.850 & 0.770 & 0.513 & 0.080 \\
\textsc{Muse}-$K$16       & 16 & 40 & 0.872 & 0.755 & 0.487 & 0.117 \\
\textsc{Muse}-$K$32       & 32 & 40 & 0.859 & 0.750 & 0.495 & 0.109 \\
\midrule
\textsc{Muse}-mt10        & 24 & 10 & 0.847 & 0.740 & 0.448 & 0.107 \\
\textsc{Muse}-mt20        & 24 & 20 & 0.859 & 0.775 & 0.508 & 0.084 \\
\textsc{Muse}-mt40        & 24 & 40 & 0.843 & 0.755 & 0.506 & 0.088 \\
\bottomrule
\end{tabular}
\end{table}

\section{Hyperparameters and Implementation}\label{app:hparams}
\Cref{tab:hparams} lists the configuration. The reflection, inner-distillation,
and answer calls all use the same frozen backbone at temperature $0$. The
mismatch control uses a fixed, seeded derangement (no fixed points) of the
meta-train user-to-support assignment. The benchmark scoring is deterministic;
no LLM judge is used. Ablation prompts are saved to label-specific sidecar files
so that the canonical $\pistar$ and its meta-test cells are never overwritten.

\begin{table}[h]
\centering
\caption{Key hyperparameters and protocol settings.}
\label{tab:hparams}
\footnotesize
\renewcommand{\arraystretch}{1.15}
\begin{tabular}{@{}l l@{}}
\toprule
\textbf{Setting} & \textbf{Value} \\
\midrule
Backbone & Qwen3-30B-A3B (frozen), 4 GPUs \\
Decoding & greedy, temperature $0$, single sample \\
Meta-train / val / test users & $40$ / $30$ / $200$ (disjoint) \\
Profile filter & $\ge 16$ profile items per user \\
Support cap $K$ (default) & $24$ (swept $\{4,8,16,32\}$) \\
Held-out query per user & $8$ (collapse recompute: $2$) \\
\textsc{Gepa} rollout budget & $120$ (default) \\
\textsc{Gepa} minibatch & $4$ \\
Reflection temperature & $0.9$ (mutation only) \\
Persona schema & summary ($\le 600$ ch.) $+ \le 6$ exemplars $+ \le 6$ rules \\
Scoring & deterministic LaMP metrics (no LLM judge) \\
Statistics & McNemar (cls.), paired bootstrap $10^4$ (reg.) \\
Seed & $0$ (item as unit of analysis) \\
\bottomrule
\end{tabular}
\end{table}

\section{Evolved Prompt Texts}\label{app:prompts}
For reproducibility we summarize the seed and evolved prompts. The seed $\pizero$
is a single free-prose instruction directing the model to read a user's labeled
history and build a persona that captures stable preferences, label-driving cues,
systematic tendencies, a few general decision rules, and representative exemplars,
emphasizing the user's \emph{decision policy} over memorization. The evolved
\Lampii{} prompt $\pistar$ reformats this into a structured JSON recipe with
fields for preferences (``extract core topics and trends\ldots focusing on
stability over short-term fluctuations''), categorization cues (``identify key
labels and surface specific cues\ldots be strict but fair''), exemplar selection,
decision rules, and tie disambiguation. The evolved \Lampiii{} prompt similarly
enumerates fields for stable preferences, noise filtering, label-specific cues,
systematic strictness/leniency, decision rules, exemplars, and near-tie
disambiguation. The mismatch-trained prompts are of the same character (generic,
benchmark-shaped recipes). Consistent with \cref{sec:analysis}, the edits are
formatting and generic guidance, not benchmark-transcending adaptive structure:
the evolved prompts cluster by benchmark (within-benchmark cosine
$0.723$/$0.729$ $>$ cross-benchmark $0.665$) rather than by whether their training
supports were aligned, and cosine$(\pizero,\pistar)$ is only $0.513$/$0.539$.

\bibliographystyle{IEEEtran}
\bibliography{references,references_extra}

\end{document}